\documentclass[letterpaper]{article} % DO NOT CHANGE THIS
\usepackage{aaai23}  % DO NOT CHANGE THIS
\usepackage{times}  % DO NOT CHANGE THIS
\usepackage{helvet}  % DO NOT CHANGE THIS
\usepackage{courier}  % DO NOT CHANGE THIS
\usepackage[hyphens]{url}  % DO NOT CHANGE THIS
\usepackage{graphicx} % DO NOT CHANGE THIS
\usepackage{natbib}  % DO NOT CHANGE THIS AND DO NOT ADD ANY OPTIONS TO IT
\usepackage{caption} % DO NOT CHANGE THIS AND DO NOT ADD ANY OPTIONS TO IT
\usepackage{algorithm}
\usepackage{algorithmic}

\usepackage{newfloat}
\usepackage{listings}
\DeclareCaptionStyle{ruled}{labelfont=normalfont,labelsep=colon,strut=off} % DO NOT CHANGE THIS
\floatstyle{ruled}
\newfloat{listing}{tb}{lst}{}
\floatname{listing}{Listing}
\usepackage{xspace}
\usepackage{url}
\usepackage{color}
\usepackage{amsmath}
\usepackage{amssymb}
\usepackage{booktabs}
\usepackage{multirow}
\newcommand{\ours}{\textsc{Aged}\xspace}

\title{Query Your Model with Definitions in FrameNet:\\ An Effective Method for Frame Semantic Role Labeling}
\author{
    Ce Zheng\textsuperscript{\rm 1},
    Yiming Wang\textsuperscript{\rm 1},
    Baobao Chang\textsuperscript{\rm 1}\thanks{\;\;Corresponding author}
}
\affiliations{
    \textsuperscript{\rm 1}The MOE Key Laboratory of Computational Linguistics, Peking University, China\\
    zce1112zslx@pku.edu.cn, w10y20ming@gmail.com, chbb@pku.edu.cn
}

\usepackage{bibentry}
\begin{document}

\maketitle

\begin{abstract}
Frame Semantic Role Labeling (FSRL) identifies arguments and labels them with frame semantic roles defined in FrameNet. Previous researches tend to divide FSRL into argument identification and role classification. Such methods usually model role classification as naive multi-class classification and treat arguments individually, which neglects label semantics and interactions between arguments and thus hindering performance and generalization of models. In this paper, we propose a query-based framework named \underline{A}r\underline{G}ument \underline{E}xtractor with \underline{D}efinitions in FrameNet (\ours) to mitigate these problems. Definitions of frames and frame elements (FEs) in FrameNet can be used to query arguments in text. Encoding text-definition pairs can guide models in learning label semantics and strengthening argument interactions. Experiments show that \ours outperforms previous state-of-the-art by up to 1.3 F1-score in two FrameNet datasets and the generalization power of \ours in zero-shot and few-shot scenarios. Our code and technical appendix is available at \url{https://github.com/PKUnlp-icler/AGED}.
\end{abstract}

\section{Introduction}

% Frame semantic parsing \cite{FSP} concentrates on extracting semantic frames as well as arguments defined in FrameNet \cite{FrameNet} from the sentence. FrameNet database is a lexical resource that contains comprehensive and hierarchical information about every frame and its frame elements (\texttt{FE}s). Extracted frame semantics can be helpful to many downstream tasks like information extraction \cite{Frame4IE}, question answering \cite{SRL4QA}, reading comprehension \cite{frame4read} and so on. This task includes three subtasks, namely Target Identification (\texttt{TI}), Frame Identification (\texttt{FI}), and Frame Semantic Role Labeling (\texttt{FSRL}). TI attempts to find predicates that might evoke frames from the sentence, and FI aims at determining frames evoked by the predicates. Based on frames identified in previous stages, FSRL tries to extract arguments that appear in the sentence and label them with frame-specific roles (i.e. FEs). We focus on FSRL in this paper.

% writefull for overleaf

Semantic Role Labeling (SRL) aims to identify arguments and label them with semantic roles for each predicate in a sentence. Frame Semantic Role Labeling (FSRL) is proposed based on frame semantics theory \citep{FSP}, where predicates are target words that can evoke semantic frames, and arguments are labeled with frame elements (FEs). Frames and FEs are described in the lexical resource FrameNet \citep{FrameNet}. Frames represent different events, relations, objects, and situations. FEs of a frame are frame-specific roles. FSRL can extract frame semantic structures from text and thus can be helpful to many downstream tasks such as information extraction \citep{Frame4IE}, question answering \citep{SRL4QA}, reading comprehension \citep{frame4read}.

Previous work tends to divide the FSRL into argument identification and role classification, and they usually identify the arguments first and then assign an FE to each argument. There are two flaws in such methods. First, interactions between arguments are either neglected \citep{frameseq2seq, LinGraph} or modeled in a sophisticated and time-consuming interaction module \citep{chen2021joint, KID} where the arguments of one frame are identified sequentially. Besides, label semantics of FEs are ignored. For regular role classifiers, representations of FEs are learned from labeled instances. However, limited by amount of training data of FSRL, such methods may perform poorly, especially for low-resource and few-shot frames because they cannot capture label semantics directly. 

In relevant tasks, e.g. SRL \citep{carreras-marquez-2005-introduction} and Event Argument Extraction (EAE) \citep{ebner-etal-2020-multi}, some researches propose query-based frameworks to handle problems mentioned above. They treat these tasks as Machine Reading Comprehension (MRC) \citep{liu-etal-2020-event} or fill-in-the-bank Clozing \citep{PAIE}. Queries are generated from role-specific questions \citep{queryner} or event (or event role) templates to extract argument spans,  where label semantics can be captured in label-specific templates. Recent work in EAE \citep{PAIE} uses event templates to extract all arguments simultaneously, where interactions between arguments are highlighted in the event templates. However, the templates or questions used by them are either too simple or in need of elaborative manual design. The simple questions format is not informative enough 
for FSRL, and manually designed templates cannot be easily applied to FrameNet because there are nearly 1000 frames and 10000 frame elements in FrameNet.

\begin{figure*}[h]
    \centering
    \includegraphics[width = 1.8\columnwidth]{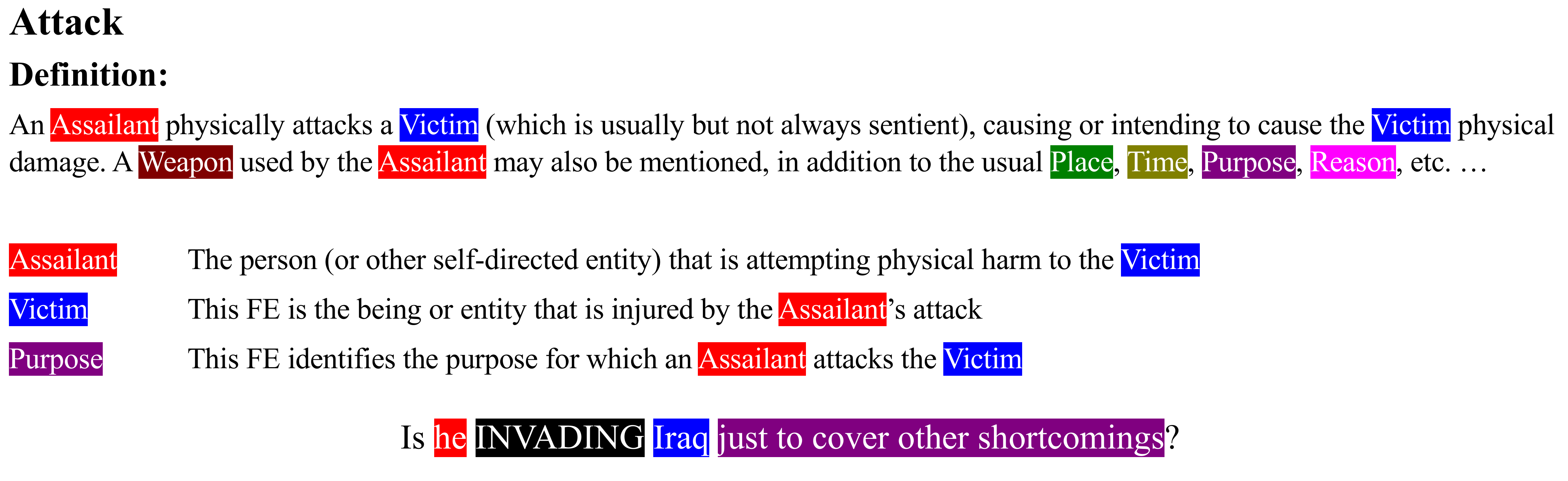}
    \caption{Definitions of \texttt{Attack} and its FEs
    \texttt{Assailant}, \texttt{Victim}, and \texttt{Purpose} with an example sentence of frame \texttt{Attack} and these FEs. \textit{INVADING} is the target word evoking frame \texttt{Attack}, arguments of the target are painted with the same color they correspond to, e.g., \textit{he} is associated with FE \texttt{Assailant}.}
    \label{fig1}
\end{figure*}

FrameNet describes frames and FEs with frame definitions and FE definitions. As shown in Figure \ref{fig1}, the frame definition of \texttt{Attack} describes an event where an \texttt{Assailant} attacks a \texttt{Victim}, and other FEs of \texttt{Attack} such as \texttt{Weapon} are also included. The definition of \texttt{Attack} also describes how they interact with each other; FE definitions show fine-grained descriptions of FEs and FE relations, for example, the definition of \texttt{Assailant} shows the relation \texttt{be\_attack\_by} between \texttt{Victim} and \texttt{Assailant} and vice versa. %Definitions in 

We find that frame definitions and FE definitions are suitable for the above problems. First, frame definitions describe how FEs of this frame interact with each other, and FE definitions also show fine-grained relations between FEs (e.g. the definition of FE \texttt{Victim} describes relations between \texttt{Victim} and \texttt{Assailant}). We can model relations between FEs with these definitions, and relations between FEs (e.g., \texttt{Assailant} and \texttt{Victim}) can reflect interactions between arguments (e.g., \textit{he} and \textit{Iraq}). Additionally, we treat definitions as natural language style queries, and representations of FEs are encoded by pretrained language models (PLMs), which can utilize label semantics of FEs, such as FE names and their context in definitions. Moreover, the definitions can be applied directly as templates with only a few modifications, which is informative and free of time-consuming manual design.

In this paper, we propose a query-based frame semantic role labeling framework named \textbf{\underline{A}}r\textbf{\underline{G}}ument \textbf{\underline{E}}xtractor with \textbf{\underline{D}}efinitions in FrameNet (\ours). We concatenate the text with frame definitions and use PLMs to encode text-definition pairs. \ours can extract all arguments simultaneously. For each FE, its representation is derived from contextual representations in PLM. Label semantics of this FE, and interactions between FEs can be captured with the bidirectional attention mechanism in PLM. Each FE representation can generate two pointer queries to identify start and end positions of arguments. In addition, we can use text-FE definition pairs as extra training data because FE definitions can represent fine-grained label semantics and FE relations.

Experiments show that \ours~outperforms previous state-of-the-art models by up to 1.3 F1-score points in two FrameNet datasets. Further experiments demonstrate the power of \ours in zero-shot and few-shot scenarios. We also combine \ours~with \citet{fido} in multitask training paradigm to explore interactions between FSRL and frame identification. 

Overall, our contribution can be summarized as follow:
\begin{itemize}
\item We propose \ours, a query-based framework to model label semantics and strengthen interactions between arguments in FSRL. Different from traditional two-stage (argument identification and role classification) methods, \ours achieve better performance on FSRL, especially in zero-shot and few-shot scenarios.

\item We use definitions in FrameNet as templates in \ours with only a few modifications. Frame definitions can be used to extract all arguments simultaneously; while FE definitions can serve as additional training data to capture fine-grained label semantics and FE relations.

\end{itemize}

\section{Task Formulation}
Frame Semantic Role Labeling aims to identify arguments and label them with frame elements for frame-evoking targets in a sentence. For a sentence $S = w_1,\dots,w_n$ and a target word $w_t$ that evokes a frame $f$. Suppose that the arguments for the predicate $w_t$ are $a_1, \dots, a_k$, and we are required to identify the start and end positions $s_i$ and $e_i$ for each argument $a_i = w_{s_i}, \dots, w_{e_i}$ and label $a_i$ with the semantic role $r_i \in \mathcal{R}_f$, where $\mathcal{R}_f$ are frame elements of the frame $f$.

Previous researches usually adopt methods including argument identification and role classification:
\begin{itemize}
    \item Argument Identification: the start and end positions ($s_i$, $e_i$) for each argument $a_i$ are identified first.
    \item Role Classification: based on $a_i$, $w_t$, and $S$, an FE $r_i \in \mathcal{R}_f$ is assigned to $a_i$.
\end{itemize}

In this work, we use definitions in FrameNet as templates and FEs in definitions as slots; thus, FSRL can be treated as slot filling. Frame definition $D_f$ of frame $f$ contains all FEs $\mathcal{R}_f=\{r_1, \dots,r_m\}$, and we need to fill these slots in $D_f$. For each slot $r_i$:
\begin{itemize}
    \item First, we need to determine whether there exists an argument $a_i$ labeled with $r_i$ in sentence $S$.
    \item If there exists an argument $a_i$ labeled with $r_i$ in sentence $S$, we need to identify $s_i$ and $e_i$ for this argument. 
\end{itemize}

\section{Methodology}
\begin{figure*}[h]
    \centering
    \includegraphics[width = 1.8\columnwidth]{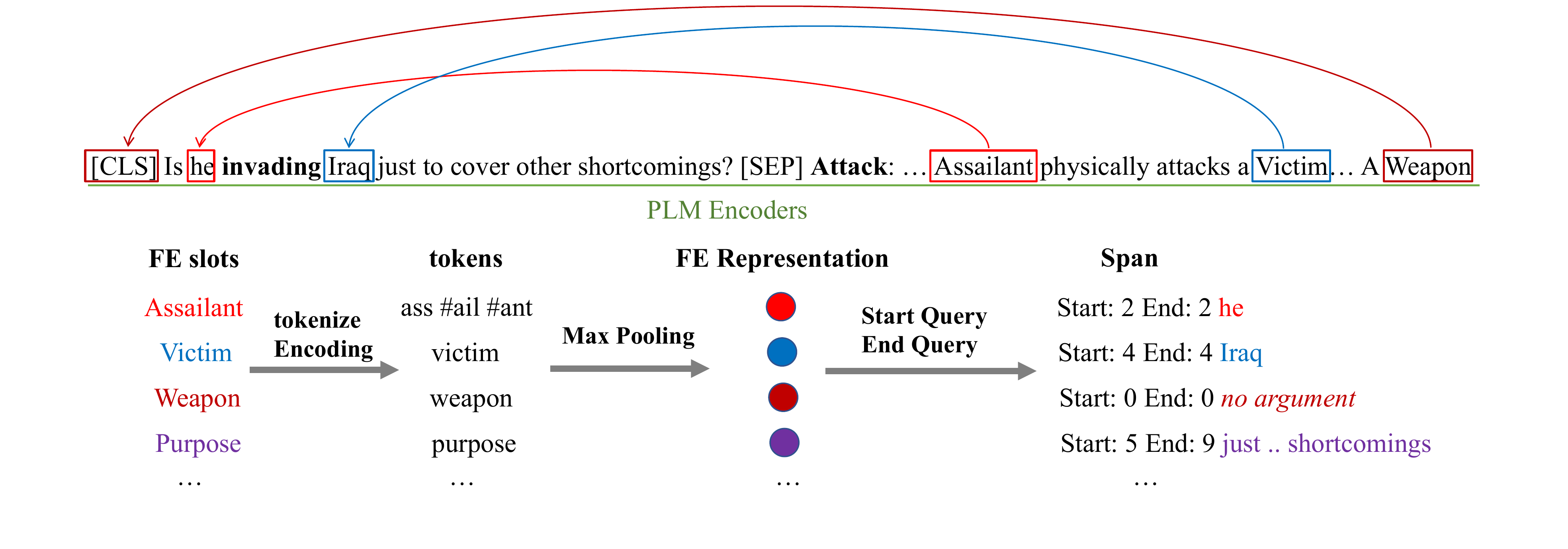}
    \caption{\ours generate queries for each FE slot in the definition and arguments are extracted based on queries.}
    \label{fig2}
\end{figure*}

We propose a query-based framework for FSRL named \ours. FSRL is modeled as template clozing. \ours utilize definitions in FrameNet as templates, and FEs in definitions as slots, then we can generate queries for each FE to extract arguments to fill the slots. Figure \ref{fig2} shows the how \ours build queries from definitions and extract arguments.

%model introduction & illustration.

\subsection{Definitions in FrameNet}
Definitions in FrameNet include frame definitions and FE definitions, which describe label semantics of frames or FEs, and semantic relations between FEs. \ours encodes text-definition pairs and both definitions can be appended to the text.
\subsubsection{Frame Definitions}
Frame definitions describe frames and their FEs in a global view. In other words, the definition of a frame describe how its FEs interact with each other to combine this frame. Frame definitions are similar to event templates in EAE, and according to \citet{PAIE}, using such templates can not only extract all arguments of an event simultaneously, but can also capture argument interactions. If we add frame definitions to the text, \ours can also extract arguments to fill all FE slots in frame definitions and capture argument interactions. However, frames in FrameNet usually contain more semantic roles than events in EAE, while some FEs are not mentioned in frame definitions, especially some non-core FEs. An easy and straightforward solution is to add other FEs not mentioned in the raw definitions to them:
\begin{equation}
    D_f = \texttt{frame name|raw def|FE list} \label{eq1}
\end{equation}
For a frame $f$, we concatenate its name \texttt{frame name}, its raw definition \texttt{raw def} and \texttt{FE list} together. For all FEs $\mathcal{R}_f$ of frame $f$, we remove FEs that have been included in \texttt{raw def}, and remaining FEs are listed in the order pre-defined by FrameNet project to construct \texttt{FE list}. Now $D_f$ ensures that any role $r \in \mathcal{R}_f$ has a slot in the definition, and \ours can extract all arguments in this frame from text with $D_f$.

\subsubsection{Frame Element Definitions}
Frame element definitions give precise and concrete descriptions of frame elements. FE definitions also show fine-grained relations between FEs and can help \ours capture argument interactions better. For example, the FE definition of \texttt{Assailant} tells the relation \texttt{be\_attack\_by} between \texttt{Victim} and \texttt{Assailant}, and reflects the semantic relation between arguments \textit{he} and \textit{Iraq}, so it is helpful to capture arguments' interactions to extract arguments. Accordingly, we also append FE definitions to text in training stage as extra training data. For a frame $f$ and an FE $r\in \mathcal{R}_f$, we concatenate its \texttt{frame name} and \texttt{FE name} and \texttt{raw def} together. Different from frame definition, we use both \texttt{frame name} and \texttt{FE name} here to distinguish FEs of different frames that have the same name; we do not add \texttt{FE list} because for FE definition, we do not aim at extracting all arguments, instead, we only care FEs that are related to current FE.
\begin{equation}
    D_r = \texttt{frame name|FE name|raw def} \label{eq2}
\end{equation}

\subsubsection{Definitions v.s. Role Specific Questions}
A common approach in query-based framework is to use simple role-specific questions like \texttt{What is the Assailant of Attack?} or \texttt{Who attacks Iraq?}. To compare the performance of definitions in FrameNet with simple questions, we also use plain question here:
\begin{equation}
    Q_r = \texttt{What's [FE name] of [frame name]?} \label{eq3}
\end{equation}
This kind of question template is simple and can be applied to any FE of any frame. However, this template only models label semantics by its name and ignores semantic relations between FEs. Moreover, a text-question pair can only extract one argument, so it will be slower than text-definition pairs.

\subsubsection{Frame definitions v.s. FE definitions}
Both the frame definitions and the FE definitions can be appended to the text to extract arguments. Frame definitions are used in both the training and inference stages because frame definitions can be efficient in extracting all arguments from a global view in one shot. FE definitions implies fine-grained label semantics of FEs and semantic relations between FEs.

 FE mentions in the definitions are slots to be filled. \footnote{As is shown in figure \ref{fig1}, there are two \texttt{Assailant} mentions in \texttt{Attack} definitions, and we only use the leftmost mention as its slot.} Slots in frame definitions include all FEs of this frame and slots in FE definitions only include this FE and its related FEs. 

\subsection{Model Architecture}
\ours, shown in Figure \ref{fig2}, uses PLMs to encode text-definition pairs to contextualized representations and generate queries for each FE slot in the definition. Argument queries can capture label semantics of FEs and semantic relations between arguments, then they are used to identify start and end positions of each arguments.
\subsubsection{Text Encoder}
Text-definition pairs are fed to the PLMs in the form of \texttt{[CLS] text [SEP] definition [SEP]}. The contextualized representations of each token in both the text and the definition are then derived from PLMs. The definition can be viewed as context of tokens in text and vice versa, thus alignments between arguments in text and FEs in the definition can be learned via self-attention mechanism of PLMs.
\begin{equation}
    H^S; H^D = \mathrm{Encoder} \left(S;D\right)
\end{equation}
$H^S = \left(h^S_0, h^S_1, \dots, h^S_n\right)$ are contextualized representations of tokens $w_1, \dots, w_n$ in sentence $S$ and $h^S_0$ are contextualized representation of token \texttt{[CLS]}. Analogously, $H^D$ are contextualized representations of tokens in definition $D$.

To make PLMs focus on targets and frame or FE mentions,  we also add special tokens in text-definition pair. \texttt{<t>} and \texttt{</t>} are inserted to the left and right of the target word $w_t$: 
\
\begin{equation}
 S=w_1, \dots, \texttt{<t>}, w_t, \texttt{</t>}, \dots, w_n   
\end{equation}

Similarly, \texttt{<f>} and \texttt{</f>} are placed around \texttt{frame name}, and \texttt{<r>} and \texttt{</r>} are placed around all FE mentions in the definition.

\subsubsection{Query Generator and Span Pointer}
Based on contextualized representations of tokens in text and definition, \ours generate queries for each FE slot in the definition, and use these queries to identify start and end positions of arguments.

FE slots are FE mentions in the definition. If an FE has more than one mentions in the definition, we choose the leftmost one as its slot. The slots in a frame definition are all FEs of this frame, while slots in an FE definition contain this FE and FEs related to it. For $m$ slots in the definition $D$, $\mathrm{slot}_1, \dots, \mathrm{slot}_m$, each slot is an FE mention span $\mathrm{slot}_j = w^{D}_{s'_j}, \dots,w^{D}_{e'_j}$, where $s'_j$ and $e'_j$ are start and end positions of $\mathrm{slot}_j$ in definition $D$. 

\begin{equation}
    q_j = \mathrm{Maxpooling} \left(h^{D}_{s'_j}, \dots,h^{D}_{e'_j} \right)
\end{equation}

$q_1, \dots, q_m$ are query vectors generated from $\mathrm{slot}_1, \dots, \mathrm{slot}_m$. Then we use pointer networks to identify start and end positions $i_s$ and $i_e$ for each slot $\mathrm{slot}_i$.

\begin{align}
    \mathrm{Pr}(s_i|S, D, q_i) &= \mathrm{Softmax}\left((W^sq_i)^\top \cdot {H^S}\right) \\
    \mathrm{Pr}(e_i|S, D, q_i) &= \mathrm{Softmax}\left((W^eq_i)^\top \cdot {H^S}\right)
\end{align}

Here $q_i \in \mathbb{R}^d$ and $H^S = (h^S_0,h^S_1, \dots, h^S_n) \in \mathbb{R}^{d\times (n+1)}$. $W^s$ and $W^e$ are linear transformation matrices in $\mathbb{R}^{d\times d}$. $\mathrm{Pr}({s_i/e_i}|S, D, q_i)$ is the probability distribution of $(n+1)$ tokens as the start or end position of $\mathrm{slot}_i$ and the token 0 (\texttt{[CLS]}) means \textit{no argument}.

\subsection{Training and Inference}
\subsubsection{Training}
In training stage we use both frame definitions and FE definitions. An instance in the training data includes a sentence $S$, and the target $w_t$ that invokes the frame $f$, and the $k$ arguments $a_1, \dots, a_k$ of the target $t$ labeled with FEs $r_1, \dots, r_k$. We use $\left(S, D_f\right)$ as original training data. Furthermore, for each $r_i$, we use $\left(S, D_{r_i}\right)$ as additional training data. The labels of each slot are the start and end positions of arguments related to slots. If there exists an argument $w_{\hat{s}_i}, \dots, w_{\hat{e}_i}$ labeled with $\mathrm{slot}_i$, the labels of this slot are $\hat{s}_i$ and $\hat{e}_i$. If there is no argument labeled $\mathrm{slot}_i$, the labels of this slot are $\hat{s}_i=0$ and $\hat{e}_i=0$ (that is, we regard \texttt{[CLS]} as no argument). We use cross entropy loss in \ours.
\begin{align}
\mathcal{L}_s &= \sum_{i=1}^{m}-\log\mathrm{Pr}(\hat{s}_i|S, D, q_i)\\
\mathcal{L}_e &= \sum_{i=1}^{m}-\log\mathrm{Pr}(\hat{e}_i|S, D, q_i)\\
\mathcal{L} &= 0.5\mathcal{L}_s + 0.5\mathcal{L}_e
\end{align}

\subsubsection{Inference}
We only use frame definitions in inference stage because a frame definition contains all FE of this frame and \ours can extract all arguments efficiently in one shot.

For each FE slot, we first assume that there exists an argument, and we adopt a greedy strategy to identify the argument span, which means the probability of an argument span $(s_i, e_i)$ labeled with $a_i$ equals $\mathrm{Pr}(s_i|S, D, q_i)\cdot\mathrm{Pr}(e_i|S, D, q_i)$:
\begin{align}
    s^{pred}_i &= \arg\max_{s_i \neq 0} \mathrm{Pr}(s_i|S, D, q_i) \\
    e^{pred}_i &= \arg\max_{e_i \neq 0} \mathrm{Pr}(e_i|S, D, q_i)
\end{align}
We also add some constraints: 
\begin{itemize}
    \item To identify a valid span, $e^{pred}_i$ should be no less than $s^{pred}_i$: $e^{pred}_i >= s^{pred}_i$.
    \item If $\mathrm{Pr}(s_i=0|S, D, q_i)\cdot\mathrm{Pr}(e_i=0|S, D, q_i)$, the probability of \textit{no argument}, is greater than $\mathrm{Pr}(s_i=s^{pred}_i|S, D, q_i)\cdot\mathrm{Pr}(e_i=e^{pred}_i|S, D, q_i)$, the predicition of this slot will be \textit{no argument}.
\end{itemize}

\section{Experiments}
We mainly focus on these questions and conduct corresponding experiments:
\begin{itemize}
    \item What is the performance of \ours in FrameNet datasets?
    \item What is the difference between definitions and simple role-specific questions?
    \item Can definitions in FrameNet help \ours capture label semantics? How does \ours perform in few-shot and zero-shot scenario?
\end{itemize}
\begin{table}
\centering\small
\begin{tabular}{lcccccc}
\toprule
        & \#frame & \# FE    & \#exem & \#train & \#dev & \#test \\ \midrule
FN 1.5 & 1019& 9634& 153952     & 17143   & 2333  & 4458\\
FN 1.7 & 1221 & 11428& 192461     & 19875   & 2309  & 6722\\ \bottomrule
\end{tabular}
\caption{Comparison between FN 1.5 and FN 1.7. FN 1.7 defines more frames and FEs than FN 1.5, and contains more training and test instances. ``exem'' means exemplar annotated sentences for frames and lexical units in the FrameNet.
}
\label{tab1}
\end{table}
\subsection{Datasets}
We evaluate the performance of \ours in two FrameNet datasets: FN 1.5 and FN 1.7. FN 1.7 is an extension version of FN 1.5. Table \ref{tab1} shows the comparison between these two datasets. The train / dev / test split is the same as \citet{frame+dep}. FrameNet also annotates exemplar sentences for frames and their lexical units, and these exemplar sentences are usually used as additional training data in previous researches \citep{frame_hetero, biLSTM+CRF+joint, frame+dep, KID}. We follow these researches and also use exemplar sentences as extra training data.

\begin{table*}[t]
\centering\small
%\resizebox{.95\columnwidth}{!}{
\begin{tabular}{lcccccc}
 \toprule
 \multirow {2} {*} {Model}& 
  \multicolumn{3}{c}{\textbf{FN 1.5}} & \multicolumn{3}{c}{\textbf{FN 1.7}} \\
                 &Precision             & Recall             & F1-score&Precision             & Recall             & F1-score\\ \midrule
semi-CRF \citeyearpar{segRNN+segmental}    &    - & -          & 73.56  & - & -& 72.22         \\
\citet{LinGraph}& - & -          & 73.28 &- & - & 72.06\\
\citet{frameseq2seq} & - & - &- & 71 & 73 & 72\\
\ours (ours) w/o exemplar & 71.93 & \textbf{76.78} & 74.28 & 74.02 & \textbf{75.46} & 74.73 \\
\ours (ours) + FE definition  w/o exemplar& \textbf{73.43} & 76.31 & \textbf{74.84}&\textbf{75.39}&75.36&\textbf{75.37}\\\midrule
\citet{chen2021joint}   &69.27&75.39&72.20&-&-&-     \\
\citet{GCN+constituency} & \textbf{74.23} & 76.94 & 75.56&-&-&- \\
\textsc{Kid} \citeyearpar{KID}  & 71.7 &79.0 & 75.2&74.1&77.3&75.6\\
\ours (ours) & 73.06 & \textbf{79.84} & 76.30 & \textbf{75.84} & 77.87 & 76.84\\
\ours (ours) + FE definition &74.04 &79.75& \textbf{76.79}& 75.80&\textbf{78.05} & \textbf{76.91}\\
\bottomrule
           
\end{tabular}
\caption{Empirical results on the test set of FN 1.5 and FN 1.7. Models in the upper block do not use exemplar instances others in the bottom block use exemplar instances as additional training data. \ours outperforms previous state-of-the-art by up to 1.3 F1 score (75.6 $\to$ 76.91). \ours still performs better than other models when trained with only frame definitions.}
\label{tab2}
\end{table*}

% \subsection{Experiment Setup}
% We use \texttt{bert-base-uncased} as the PLM in \ours. We follow \citet{chen2021joint, KID} to first train \ours on exemplar sentences then train on the train set continually. We search for hyperparameters (learning rate, batch size, and epoch num) with performance in the development set. Performance in the development set is also used to save the best parameters of the models, and we will evaluate \ours with these parameters in the test set. 

% Our code is implemented with \texttt{Pytorch} and \texttt{Huggingface}. \ours is trained on \texttt{NVIDIA A40} with 40 GB memory and it will take about 4 GPU hours to train \ours and 0.6 hours when \ours is trained only with the training dataset. More details about hyperparameter settings are listed in the Appendix.

% The performance of \ours is evaluated by the micro-F1 score\footnote{\url{https://www.cs.cmu.edu/~ark/SEMAFOR/eval/}}. For an argument span, we consider that the exact match that requires the prediction triple \texttt{(FE, start, end)} should be the same as the ground truth.
\subsection{Main Results}
We compare \ours with previous models for both FN 1.5 and FN 1.7. For fair comparison, we only compare the performance of \ours with models that use PLMs\footnote{Experiment setups and details of these models are listed in \url{https://github.com/PKUnlp-icler/AGED}}. These models can be roughly divided into two groups: w/o exemplar and w/ exemplar because some models claim that they do not use exemplar instances as additional training data. 

Table \ref{tab2} shows the main results of our experiments. If \ours is trained with only frame definitions, \ours outperforms the previous state of the art by up to 1.2 F1-score (75.6 $\to$ 76.84). Using FE definitions as extra training data can help \ours performs better and it will outperform the previous state of the art by up to 1.3 F1-score (75.6 $\to$ 76.91). The results show the effectiveness of \ours under all conditions in both datasets, which indicates that the models can benefit from considering label semantics and argument interactions with definitions in FrameNet. Using FE definitions as extra training data can bring up to 0.6 F1-score increase (74.73 $\to$ 75.37) because these definitions give concrete and precise descriptions of FEs and reflect fine-grained FE semantic relations.

\subsection{Discussion}
In this section, we conduct further experiments for better understanding of \ours. If not specified, experiments is conducted on FN 1.5 without exemplar sentences because training with exemplar sentences is time-consuming.

\subsubsection{Few-shot and Zero-shot Performance}
\citet{KID} design a few/zero-shot experiment to validate the transfer learning ability of the Frame Knowledge Graph constructed by them. We follow the same experiment settings to evaluate \ours with three frames \texttt{Getting}, \texttt{Arriving}, \texttt{Transition\_to\_state} under zero-shot and few-shot scenarios. Table \ref{tab3} shows the comparison between \ours and \textsc{Kid} under zero-shot and few-shot scenarios. As they do not use PLMs in this experiment, we cannot directly compare zero-shot results of two models (56.25 v.s. 82.45). When removing all instances of these frames, \textsc{Kid} drops much larger than \ours (-14.06 v.s. -3.81), and the situations is the same under few-shot scenarios (full $\to$ 32). Definitions can help \ours directly capture label semantics, while \textsc{Kid} uses semantic relations between Frames and FE mappings in semantic-related frames to transfer knowledge. Results show that using definitions directly is much more effective under zero-shot and few-shot scenarios. Even \ours never see instances of these frames in training stage, its performance is still close to \ours with full instances because \ours learns ability to parse definitions as queries in training stage.

As FN 1.7 is an extension version of FN 1.5, we also conduct another zero-shot experiment for \ours. FN 1.7 defines some new frames which are not included in FN 1.5 and we wonder the performance of \ours trained with FN 1.5 on FN 1.7. Table \ref{tab1} shows that FN 1.7 defines 202 new frames and the test dataset of FN 1.7 includes at least 2264 new instances. Nearly 25\% of these 2264 new instances evoke new frames in FN 1.7. Table \ref{tab4} shows the results of this experiment. As there are some new frames, \ours trained with FN 1.5 performs slightly worse than \ours trained with FN 1.7 (73.72 v.s. 74.73). The narrow gap indicates the zero-shot performance of \ours. To further validate the performance of \ours with unseen frames, we evaluate \ours trained with FN 1.5 on instances of new frames in FN 1.7 test set, and the gap is still narrow (72.86 v.s. 74.73). 

The two above experiments both show the generalizability of \ours. Even if these frames are never or seldom seen in training stage, \ours can still perform promising if the corresponding definitions are provided and the model learns how to encode them.

\begin{table}[]
    \centering\small
    \begin{tabular}{lccc}
    \toprule
    \multirow{2}{*}{Model} & \multicolumn{3}{c}{$K$}  \\ \cmidrule{2-4}
    & 0  & 32 & full\\
    \midrule
         \textsc{Kid} (GloVe) & 56.26 & 65.95 & 70.32 \\
         \ours & 82.45 & 84.06 & 86.26\\
    \bottomrule
    \end{tabular}
    \caption{Experiments under zero-shot and few-shot scenarios. Experiment settings is from \citet{KID}. In zero-shot scenarios, \ours performs much promising in zero-shot and few-shot scenarios than \textsc{KID} \citep{KID}. The difference between zero-shot and full instances is quite small (82.45 $\to$ 86.26), which verifies that definitions can bring the power of generalization and transfer learning.}
    \label{tab3}
\end{table}

\begin{table}[t]
\centering\small
%\resizebox{.95\columnwidth}{!}{
\begin{tabular}{lccccc}
 \toprule
 Model & Precision             & Recall             & F1-score & $\Delta$\\\midrule
15 $\to$ 17 &73.28&74.17&73.72 & -1.01 \\
15 $\to$ 17 (new) & 71.85 & 73.90 & 72.86 & -1.87 \\
17 $\to$ 17 &74.02 & 75.46 & 74.73 & - \\ 

\bottomrule
           
\end{tabular}
\caption{Zero-shot experiments from FN 1.5 to FN 1.7. 15 $\to$ 17 means training with FN 1.5 and evaluating with FN 1.7. 15 $\to$ 17 (new) means evaluating with instances of new frames in FN 1.7. The F1-score of 15 $\to$ 17 (new) shows the zero-shot performance of \ours for new frames in FN 1.7.}
\label{tab4}
\end{table}

\begin{table}[t]
\centering\small
%\resizebox{.95\columnwidth}{!}{
\begin{tabular}{lccccc}
 \toprule
 Model & Precision             & Recall             & F1-score & $\Delta$\\\midrule
\ours&71.93 & 76.78 & 74.28 & -  \\
\ours w/o label &72.09& 76.48&74.22&-0.06 \\
\ours w/o target & 62.36 & 65.18 & 63.74 & -10.54 \\ 

\bottomrule
           
\end{tabular}
\caption{Ablation study of target markers and label markers in \ours.}
\label{tab5}
\end{table}

\begin{table}[t]
\centering\small
%\resizebox{.95\columnwidth}{!}{
\begin{tabular}{lcccc}
 \toprule
 Model & Precision             & Recall             & F1 & \#pairs\\\midrule
Role-specific QA &72.32&75.12&73.69&44252\\
\ours (FE Def) &72.53&76.00&74.23&44252\\
\ours (Frame Def) & 71.93 & 76.78 & 74.28 & \textbf{4458} \\
% \ours (All Def -) & 73.11 & 76.41 & 74.73 & \textbf{4458} \\

\ours (All Def) & 73.43 & 76.31 & \textbf{74.84} & \textbf{4458} \\\bottomrule
           
\end{tabular}
\caption{Comparison between Definitions and Role-Specific Questions. All Def means using FE definitions as additional training data. Using definitions is more effective than simple questions. Frame Def is more efficient than FE Def because it can extract all arguments simultaneously. All Def is a good way to combine frame Def and FE Def together.}
\label{tab6}
\end{table}

\begin{table*}[t]
\centering \small
%\resizebox{.95\columnwidth}{!}{
\begin{tabular}{lcccccc}
 \toprule
\multirow{2}{*}{Model} & \multicolumn{2}{c}{FI} & \multicolumn{4}{c}{FSRL}\\ & Accuracy &$\Delta$ & Precision & Recall & F1-score &$\Delta$\\\midrule
\citet{GCN+constituency} & 89.83 & - & 74.23& 76.94 &75.56 & -\\
\citet{GCN+constituency} (m) & 90.10 & +0.27 & 74.56& 74.43& 74.50 & -1.06 \\
\citet{LinGraph} & 90.16 & - & - & - & 73.56 & - \\
\citet{LinGraph} (m) & 90.62 & +0.46 & - & - & 73.22 & -0.34 \\
\midrule
\ours &90.78&-&71.93&76.78&74.28&-\\
\ours (m)&91.63&+0.85&72.18 & 76.89 & 74.46&+0.18\\
\bottomrule
           
\end{tabular}
\caption{\ours trained with single-task v.s. multi-task (m). \ours for FI is a simple reimplementation version of \citet{fido}. Unlike previous multi-task frameworks, \ours can benefit from multi-task in both subtasks especially FSRL. }
\label{tab7}
\end{table*}

\subsubsection{Target Markers and Label Markers}
A common practice to strengthen contextualized span representations is the use of markers \citep{baldini-soares-etal-2019-matching, xiao-etal-2020-denoising}. Table~\ref{tab5} gives an ablation study of markers in the text-definition pairs. Results show that label markers such as \texttt{<r>} and \texttt{<f>} bring negligible improvements because FEs are already natural language labels instead of abstract labels in Propbank (arg0, arg1, ..., ) and markers seem redundant to capture label semantics of FEs in FrameNet. However, target markers play a vital role in \ours. When we remove target markers, the performance will drop by 10.54 points. Frame semantics is based on predicate-argument semantic structure, and target (predicate) is central in frame semantics. Without target markers, \ours even does not know which word is the target, thus affecting the performance of \ours.

\subsubsection{Combination of Frame Definitions and FE Definitions}
We can use definitions for \ours and we can still use simple questions, what is the difference between simple questions and definitions? Both frame definitions and FE definitions can be used in \ours, why choose frame definitions in the inference stage and why choose FE definitions as additional training data? This section answers these questions.

In this section, we use three baselines:
\begin{itemize}
    \item Role-Specific QA: we use role specific questions in Eq.~\ref{eq3}. A frame that includes $m$ FEs will construct $m$ text-question pairs. Each pair only queries one argument.
    \item FE Def: using FE Definitions definied in Eq.~\ref{eq2} instead of simple questions. A frame including $m$ FEs will construct $m$ text-definition pairs. Each text-definition pair only queries one argument.
    \item Frame Def: baseline model in our work, using frame definitions to query all arguments with only one single text-definition pair.
\end{itemize}

Results are listed in Table \ref{tab6}. Role-specific simple questions are not informative enough than definitions in FrameNet. Using Frame Def is more efficient than FE Def as a frame includes 10 FEs on average and it will take 10x time to extract arguments when using FE Def. Besides, FE Def gets higher precision, and Frame Def gets higher recall because FE Def gives more detailed description of FEs and Frame Def extracts all arguments in a more global view. A straightforward method to combine strengths of two methods is to use FE Def as extra training data and this method achieves fast and accurate performance.

\subsubsection{Multi-Task Learning}
Frame Identification (FI) and Frame Semantic Role Labeling (FSRL) are both subtasks of Frame Semantic Parsing. Previous research \citep{GCN+constituency, chen2021joint, LinGraph, KID} has trained their models with these subtasks in an end-to-end framework. Holding the believe that interactions between subtasks can contribute to all subtasks, they usually train their models in multi-task learning. However, as reported in \citet{GCN+constituency, LinGraph}, multi-task learning is not beneficial for FSRL which means training with FSRL only performs better than multi-task.

\citet{fido} is similar to \ours because we both feed text-definition pairs to PLMs. We simply re-implement their work by removing lexical unit definitions and model FI as sentence pair multiple choice. We train \ours in both subtasks and also train \ours in multi-task learning. The results are in Table \ref{tab7}. The results of FI are consistent. All frameworks get higher performance than single-task when trained with FI and FSRL. \citet{GCN+constituency, LinGraph} cannot benefit from multi-task in FSRL while \ours trained in multi-task gets a 0.18 points improvement.

The results indicate that the use of definitions can narrow the gap between FI and FSRL, because both \citet{fido} and \ours require PLMs to understand the definitions and encode text-definition pairs. Future work on frame semantic parsing can focus on an end-to-end model with definitions.

\section{Related Work}
\noindent \textbf{Frame Semantic Role Labeling} Previous researches on FSRL \citep{frame_hetero, segRNN+segmental,  GCN+constituency, KID} range from traditional SVM classifiers to deep neural network architectures like LSTM, GCN, and pretrained language models like BERT. However, they all first find candidate argument spans and then classify them into FEs. Arguments are typically identified by sequence labeling architectures \citep{segRNN+segmental, GCN+constituency} or span-based methods \citep{biLSTM+CRF+joint, frame+dep} while interactions between arguments are neglected. Recent researches \citep{chen2021joint, KID} propose well-designed architectures to highlight interactions between arguments by identifying the arguments sequentially. Such methods implicitly consider relations between FEs and are not efficient because of their sequential modeling. Label semantics is also ignored in a typical role classifier, except \citet{KID}, where the ontology frame knowledge graph is used in their work to model structure information between labels.  

\ours directly models label semantics and interactions between arguments by encoding text-definition pairs with PLMs so it achieves fast and accurate performance.

\noindent \textbf{Definitions in FrameNet} Definitions in FrameNet are recently studied and used for Frame Semantic Parsing. \citet{fido, def+rel4FI} focus on Frame Identification. \citet{fido}  traverses all candidate frames for the same target in a sentence and appends lexical unit definition, frame definition to original sentence for each pair, then uses BERT to encode the inputs for further classification. \citet{def+rel4FI} combine frame definition with frame-to-frame relations, and they use frozen BERT to encode frame definition as node features in the frame graph. Besides, \citet{KID} extract FE relations from FE definitions to construct a frame ontological knowledge graph while the definitions are not encoded. Different from \citet{fido}, \ours concentrates on FSRL, but \citet{fido} and \ours use similar input format. 
A natural idea is to combine two models in multi-task learning which can explore interactions between these two tasks.

\noindent \textbf{Query-based Framework} Query-based methods \citep{fitzgerald-etal-2018-large, queryner, EAE-QA, qaner} are common in many NLP tasks like Semantic Role Labeling (SRL), Name Entity Recognition (NER) and Event Argument Extraction (EAE). Query-based frameworks generate queries from natural language questions or templates, and these queries are used to extract argument spans from the text that can answer the given questions or fill the slots in template clozing \citep{fitzgerald-etal-2018-large, qaner}. These frameworks show a significant improvement in the generalization of models because label semantics is contained in templates or questions. These templates are either too naive to be informative enough or in need of time-consuming human design. In FSRL, we do not need to worry about this issue because the definitions in FrameNet can directly serve as templates. 

\citet{PAIE} use event templates containing multiple roles and argue that the event templates can be efficient in extracting arguments and aware of argument interactions. Frame definitions in \ours can also extract all arguments in a frame and strengthen relations between them. Besides, we use FE definitions as additional training data, because FE definitions show fine-grained relations between FEs and relations between FEs can reflect argument interactions.

\section{Conclusion}
In this paper, we propose a query-based framework \ours for frame semantic role labeling. Frame definitions and FE definitions can capture label semantics of FEs. Semantic relations between FEs are also included in these definitions. Under the guidance of definitions, \ours achieves fast and accurate performance in two FrameNet datasets. In addition, \ours also shows the strong power of generalization for zero-shot or few-shot frames, which verifies that the label semantics is captured in \ours. Definitions in FrameNet are still potential, and further work can focus on a definition-based end-to-end framework for frame semantic parsing.

\section{Acknowledgments}
This paper is supported by the National Science Foundation of China under Grant No.61936012 and the National
Key Research and Development Program of China under Grant No. 2020AAA0106700.

\bibliography{aaai23}

\begin{thebibliography}{26}
\providecommand{\natexlab}[1]{#1}

\bibitem[{Baker, Fillmore, and Lowe(1998)}]{FrameNet}
Baker, C.~F.; Fillmore, C.~J.; and Lowe, J.~B. 1998.
\newblock The {B}erkeley {F}rame{N}et Project.
\newblock In \emph{36th Annual Meeting of the Association for Computational
  Linguistics and 17th International Conference on Computational Linguistics,
  Volume 1}, 86--90. Montreal, Quebec, Canada: Association for Computational
  Linguistics.

\bibitem[{Baldini~Soares et~al.(2019)Baldini~Soares, FitzGerald, Ling, and
  Kwiatkowski}]{baldini-soares-etal-2019-matching}
Baldini~Soares, L.; FitzGerald, N.; Ling, J.; and Kwiatkowski, T. 2019.
\newblock Matching the Blanks: Distributional Similarity for Relation Learning.
\newblock In \emph{Proceedings of the 57th Annual Meeting of the Association
  for Computational Linguistics}, 2895--2905. Florence, Italy: Association for
  Computational Linguistics.

\bibitem[{Bastianelli, Vanzo, and Lemon(2020)}]{GCN+constituency}
Bastianelli, E.; Vanzo, A.; and Lemon, O. 2020.
\newblock Encoding Syntactic Constituency Paths for Frame-Semantic Parsing with
  Graph Convolutional Networks.
\newblock \emph{CoRR}, abs/2011.13210.

\bibitem[{Carreras and M{\`a}rquez(2005)}]{carreras-marquez-2005-introduction}
Carreras, X.; and M{\`a}rquez, L. 2005.
\newblock Introduction to the {C}o{NLL}-2005 Shared Task: Semantic Role
  Labeling.
\newblock In \emph{Proceedings of the Ninth Conference on Computational Natural
  Language Learning ({C}o{NLL}-2005)}, 152--164. Ann Arbor, Michigan:
  Association for Computational Linguistics.

\bibitem[{Chen, Zheng, and Chang(2021)}]{chen2021joint}
Chen, X.; Zheng, C.; and Chang, B. 2021.
\newblock Joint Multi-Decoder Framework with Hierarchical Pointer Network for
  Frame Semantic Parsing.
\newblock In \emph{Findings of the Association for Computational Linguistics:
  ACL-IJCNLP 2021}, 2570--2578. Online: Association for Computational
  Linguistics.

\bibitem[{Du and Cardie(2020)}]{EAE-QA}
Du, X.; and Cardie, C. 2020.
\newblock Event Extraction by Answering (Almost) Natural Questions.
\newblock In \emph{Proceedings of the 2020 Conference on Empirical Methods in
  Natural Language Processing (EMNLP)}, 671--683. Online: Association for
  Computational Linguistics.

\bibitem[{Ebner et~al.(2020)Ebner, Xia, Culkin, Rawlins, and
  Van~Durme}]{ebner-etal-2020-multi}
Ebner, S.; Xia, P.; Culkin, R.; Rawlins, K.; and Van~Durme, B. 2020.
\newblock Multi-Sentence Argument Linking.
\newblock In \emph{Proceedings of the 58th Annual Meeting of the Association
  for Computational Linguistics}, 8057--8077. Online: Association for
  Computational Linguistics.

\bibitem[{FitzGerald et~al.(2018)FitzGerald, Michael, He, and
  Zettlemoyer}]{fitzgerald-etal-2018-large}
FitzGerald, N.; Michael, J.; He, L.; and Zettlemoyer, L. 2018.
\newblock Large-Scale {QA}-{SRL} Parsing.
\newblock In \emph{Proceedings of the 56th Annual Meeting of the Association
  for Computational Linguistics (Volume 1: Long Papers)}, 2051--2060.
  Melbourne, Australia: Association for Computational Linguistics.

\bibitem[{Gildea and Jurafsky(2000)}]{FSP}
Gildea, D.; and Jurafsky, D. 2000.
\newblock Automatic Labeling of Semantic Roles.
\newblock In \emph{Proceedings of the 38th Annual Meeting of the Association
  for Computational Linguistics}, 512--520. Hong Kong: Association for
  Computational Linguistics.

\bibitem[{Guo et~al.(2020)Guo, Li, Tan, Li, Guan, Zhao, and Zhang}]{frame4read}
Guo, S.; Li, R.; Tan, H.; Li, X.; Guan, Y.; Zhao, H.; and Zhang, Y. 2020.
\newblock A Frame-based Sentence Representation for Machine Reading
  Comprehension.
\newblock In \emph{Proceedings of the 58th Annual Meeting of the Association
  for Computational Linguistics}, 891--896. Online: Association for
  Computational Linguistics.

\bibitem[{Jiang and Riloff(2021)}]{fido}
Jiang, T.; and Riloff, E. 2021.
\newblock Exploiting Definitions for Frame Identification.
\newblock In \emph{Proceedings of the 16th Conference of the European Chapter
  of the Association for Computational Linguistics: Main Volume}, 2429--2434.
  Online: Association for Computational Linguistics.

\bibitem[{Kalyanpur et~al.(2020)Kalyanpur, Biran, Breloff, Chu{-}Carroll,
  Diertani, Rambow, and Sammons}]{frameseq2seq}
Kalyanpur, A.; Biran, O.; Breloff, T.; Chu{-}Carroll, J.; Diertani, A.; Rambow,
  O.; and Sammons, M. 2020.
\newblock Open-Domain Frame Semantic Parsing Using Transformers.
\newblock \emph{CoRR}, abs/2010.10998.

\bibitem[{Kshirsagar et~al.(2015)Kshirsagar, Thomson, Schneider, Carbonell,
  Smith, and Dyer}]{frame_hetero}
Kshirsagar, M.; Thomson, S.; Schneider, N.; Carbonell, J.; Smith, N.~A.; and
  Dyer, C. 2015.
\newblock Frame-Semantic Role Labeling with Heterogeneous Annotations.
\newblock In \emph{Proceedings of the 53rd Annual Meeting of the Association
  for Computational Linguistics and the 7th International Joint Conference on
  Natural Language Processing (Volume 2: Short Papers)}, 218--224. Beijing,
  China: Association for Computational Linguistics.

\bibitem[{Lin, Sun, and Zhang(2021)}]{LinGraph}
Lin, Z.; Sun, Y.; and Zhang, M. 2021.
\newblock A Graph-Based Neural Model for End-to-End Frame Semantic Parsing.
\newblock In Moens, M.; Huang, X.; Specia, L.; and Yih, S.~W., eds.,
  \emph{Proceedings of the 2021 Conference on Empirical Methods in Natural
  Language Processing, {EMNLP} 2021, Virtual Event / Punta Cana, Dominican
  Republic, 7-11 November, 2021}, 3864--3874. Association for Computational
  Linguistics.

\bibitem[{Liu et~al.(2022)Liu, Xiao, Zhu, Zhang, Li, and Arnold}]{qaner}
Liu, A.~T.; Xiao, W.; Zhu, H.; Zhang, D.; Li, S.; and Arnold, A.~O. 2022.
\newblock QaNER: Prompting Question Answering Models for Few-shot Named Entity
  Recognition.
\newblock \emph{CoRR}, abs/2203.01543.

\bibitem[{Liu et~al.(2020)Liu, Chen, Liu, Bi, and Liu}]{liu-etal-2020-event}
Liu, J.; Chen, Y.; Liu, K.; Bi, W.; and Liu, X. 2020.
\newblock Event Extraction as Machine Reading Comprehension.
\newblock In \emph{Proceedings of the 2020 Conference on Empirical Methods in
  Natural Language Processing (EMNLP)}, 1641--1651. Online: Association for
  Computational Linguistics.

\bibitem[{Ma et~al.(2022)Ma, Wang, Cao, Li, Chen, Wang, and Shao}]{PAIE}
Ma, Y.; Wang, Z.; Cao, Y.; Li, M.; Chen, M.; Wang, K.; and Shao, J. 2022.
\newblock Prompt for Extraction? {PAIE:} Prompting Argument Interaction for
  Event Argument Extraction.
\newblock In Muresan, S.; Nakov, P.; and Villavicencio, A., eds.,
  \emph{Proceedings of the 60th Annual Meeting of the Association for
  Computational Linguistics (Volume 1: Long Papers), {ACL} 2022, Dublin,
  Ireland, May 22-27, 2022}, 6759--6774. Association for Computational
  Linguistics.

\bibitem[{Meng et~al.(2019)Meng, Li, Sun, and Li}]{queryner}
Meng, Y.; Li, X.; Sun, Z.; and Li, J. 2019.
\newblock Query-Based Named Entity Recognition.
\newblock \emph{CoRR}, abs/1908.09138.

\bibitem[{Peng et~al.(2018)Peng, Thomson, Swayamdipta, and Smith}]{frame+dep}
Peng, H.; Thomson, S.; Swayamdipta, S.; and Smith, N.~A. 2018.
\newblock Learning Joint Semantic Parsers from Disjoint Data.
\newblock In \emph{Proceedings of the 2018 Conference of the North {A}merican
  Chapter of the Association for Computational Linguistics: Human Language
  Technologies, Volume 1 (Long Papers)}, 1492--1502. New Orleans, Louisiana:
  Association for Computational Linguistics.

\bibitem[{Shen and Lapata(2007)}]{SRL4QA}
Shen, D.; and Lapata, M. 2007.
\newblock Using Semantic Roles to Improve Question Answering.
\newblock In \emph{Proceedings of the 2007 Joint Conference on Empirical
  Methods in Natural Language Processing and Computational Natural Language
  Learning ({EMNLP}-{C}o{NLL})}, 12--21. Prague, Czech Republic: Association
  for Computational Linguistics.

\bibitem[{Su et~al.(2021)Su, Li, Li, Pan, Zhang, Chai, and Han}]{def+rel4FI}
Su, X.; Li, R.; Li, X.; Pan, J.~Z.; Zhang, H.; Chai, Q.; and Han, X. 2021.
\newblock A Knowledge-Guided Framework for Frame Identification.
\newblock In \emph{Proceedings of the 59th Annual Meeting of the Association
  for Computational Linguistics and the 11th International Joint Conference on
  Natural Language Processing (Volume 1: Long Papers)}, 5230--5240. Online:
  Association for Computational Linguistics.

\bibitem[{Surdeanu et~al.(2003)Surdeanu, Harabagiu, Williams, and
  Aarseth}]{Frame4IE}
Surdeanu, M.; Harabagiu, S.; Williams, J.; and Aarseth, P. 2003.
\newblock Using Predicate-Argument Structures for Information Extraction.
\newblock In \emph{Proceedings of the 41st Annual Meeting of the Association
  for Computational Linguistics}, 8--15. Sapporo, Japan: Association for
  Computational Linguistics.

\bibitem[{Swayamdipta et~al.(2017)Swayamdipta, Thomson, Dyer, and
  Smith}]{segRNN+segmental}
Swayamdipta, S.; Thomson, S.; Dyer, C.; and Smith, N.~A. 2017.
\newblock Frame-Semantic Parsing with Softmax-Margin Segmental RNNs and a
  Syntactic Scaffold.
\newblock \emph{CoRR}, abs/1706.09528.

\bibitem[{Xiao et~al.(2020)Xiao, Yao, Xie, Han, Liu, Sun, Lin, and
  Lin}]{xiao-etal-2020-denoising}
Xiao, C.; Yao, Y.; Xie, R.; Han, X.; Liu, Z.; Sun, M.; Lin, F.; and Lin, L.
  2020.
\newblock Denoising Relation Extraction from Document-level Distant
  Supervision.
\newblock In \emph{Proceedings of the 2020 Conference on Empirical Methods in
  Natural Language Processing (EMNLP)}, 3683--3688. Online: Association for
  Computational Linguistics.

\bibitem[{Yang and Mitchell(2017)}]{biLSTM+CRF+joint}
Yang, B.; and Mitchell, T. 2017.
\newblock A Joint Sequential and Relational Model for Frame-Semantic Parsing.
\newblock In \emph{Proceedings of the 2017 Conference on Empirical Methods in
  Natural Language Processing}, 1247--1256. Copenhagen, Denmark: Association
  for Computational Linguistics.

\bibitem[{Zheng et~al.(2022)Zheng, Chen, Xu, and Chang}]{KID}
Zheng, C.; Chen, X.; Xu, R.; and Chang, B. 2022.
\newblock A Double-Graph Based Framework for Frame Semantic Parsing.
\newblock \emph{CoRR}, abs/2206.09158.

\end{thebibliography}

\end{document}